\documentclass[11pt]{article}
\usepackage[a4paper,margin=1in]{geometry}

\usepackage{graphicx}
\usepackage{multirow}
\usepackage{amsmath,amssymb,amsfonts}
\usepackage{mathrsfs}
\usepackage{xcolor}
\usepackage{textcomp}
\usepackage{booktabs}

  \usepackage[authoryear,round]{natbib}
  \usepackage{hyperref}
  \hypersetup{
    pdftitle={Aligning Validation with Deployment in Spatial Prediction: Target-Weighted Cross-Validation},
    pdfauthor={Alexander Brenning, Thomas Suesse}
}

\title{Aligning Validation with Deployment in Spatial Prediction: Target-Weighted Cross-Validation}

\author{%
Alexander Brenning$^{1,2}$\thanks{Correspondence: Alexander Brenning (\texttt{alexander.brenning@uni-jena.de})}%
\and
Thomas Suesse$^{1,2}$%
}

\date{}

\begin{document}


\maketitle

\noindent
$^1$ Friedrich Schiller University Jena, Department of Geography, Jena, Germany \\
$^2$ ELLIS Unit Jena, Jena, Germany

\begin{abstract}
Reliable estimation of predictive performance is essential for spatial environmental modeling, where machine-learning models are commonly used to generate spatially continuous maps from unevenly distributed observations. Standard cross-validation (CV) procedures implicitly assume that validation data are representative of prediction conditions across the target domain. In practice, this assumption is often violated due to preferential or clustered sampling designs, leading to biased estimates of predictive performance and associated uncertainty.

We introduce a deployment-oriented validation framework for spatial prediction based on weighted CV, which aims to align validation tasks with the distribution of prediction tasks across a specified target domain. The framework encompasses importance-weighted cross-validation (IWCV) as a general solution, and a calibration-based formulation, Target-Weighted Cross-Validation (TWCV), that leverages spatially meaningful task descriptors, including environmental covariates and prediction distance.

Through controlled simulation experiments, we show that conventional non-spatial and spatial CV strategies can exhibit substantial bias under realistic sampling designs, whereas the proposed weighted CV approaches with spatial task descriptors nearly eliminate or substantially reduce this bias when validation tasks adequately cover the deployment-task space. A case study on mapping nitrogen dioxide (NO$_2$) concentrations across Germany demonstrates that standard CV approaches can overestimate prediction error due to sampling biases, while weighted CV yields estimates that are more consistent with deployment conditions.

The proposed framework separates validation task generation from risk estimation and can provide approximately unbiased estimators of deployment risk under appropriate coverage and weighting conditions. It provides a practical approach for improving the reliability of performance assessment in environmental mapping applications where sample distributions differ from prediction domains.
\end{abstract}


\section{Introduction}\label{sec:introduction}

Reliable estimation of predictive performance is central to machine learning workflows. In many applications, including environmental mapping, model selection and assessment rely on cross-validation (CV) when independent test data are unavailable \citep{pohjankukka2017,roberts2017,schratz2019,ploton.et.al.2020}. Standard CV implicitly assumes that validation observations are representative of prediction tasks encountered during deployment. Under this assumption, CV provides an approximately unbiased estimator of predictive risk \citep{hastie2009elements,arlot2010survey}.

In practice, this assumption is often violated because the distribution of prediction tasks differs from that induced by the sampling design. Such situations are commonly described as \emph{dataset shift} or \emph{covariate shift} \citep{quinonero2009dataset,sugiyama2007covariate}. When deployment and validation distributions differ, CV may estimate performance under conditions that do not reflect deployment tasks. Spatial prediction provides a particularly clear example: environmental monitoring networks are often preferentially located, whereas predictions are required across an entire geographic domain. Consequently, CV evaluates tasks defined by the monitoring network rather than those across the full prediction domain \citep{Wadoux2021}.

Beyond covariate shift, spatial prediction also highlights \emph{task-difficulty shift}. Prediction distance is a key driver of prediction skill under spatial dependence \citep{brenning.2023.ijgis}, and differences in prediction-distance distributions between validation and deployment can strongly affect performance estimates, especially under clustered sampling \citep{debruin2022,mila.et.al.2023}. In this context, CV can be viewed as a two-step process: (i) a \emph{task generator} that samples validation tasks, and (ii) a \emph{risk estimator} that aggregates their losses. Under dataset shift, bias may arise from either component, a distinction formalized in Sect.~\ref{sec:deploymentrisk}.

Previous work has shown that spatial dependence between training and validation samples can lead to optimistic performance estimates \citep{roberts2017,karasiak.et.al.2021}. Spatially structured resampling therefore aims to reduce dependence by enforcing spatial separation \citep{brenning2005,brenning.2012.sperrorest,roberts2017,schratz2024jss}. Related approaches such as nearest-neighbour distance matching attempt to approximate deployment conditions through resampling \citep{mila.et.al.2023,linnenbrink.2023.nndmcv}, while other work proposes diagnostics of extrapolation in feature space \citep{meyer.pebesma.2021.aoa,ploton.et.al.2020} or distance-dependent performance \citep{brenning.2023.ijgis}. However, these approaches either modify validation design or provide diagnostics rather than directly targeting predictive risk under the deployment-task distribution.

This suggests that aligning validation with deployment conditions requires addressing distributional mismatch at the level of the risk estimator. We propose \emph{Target-Weighted Cross-Validation} (TWCV), a weighted CV estimator targeting predictive performance under the distribution of tasks across a specified deployment domain. TWCV can be interpreted as a calibration-based approximation to importance-weighted risk estimation \citep{shimodaira2000improving,sugiyama2007covariate}, in which weights are constructed by matching marginal distributions of task descriptors rather than estimating density ratios. It assigns weights so that the weighted distribution of validation tasks matches the covariate and task-difficulty distributions across the target domain. The weighting scheme is obtained using calibration weighting (raking) \citep{deville1992calibration,lumley.2004.jss}. By incorporating prediction distance as a task descriptor \citep{brenning.2023.ijgis}, TWCV accounts for both covariate and task-difficulty shift through reweighting. Unlike approaches that are limited to modifying the resampling design, TWCV operates at the level of the risk estimator.

This paper makes three main contributions:
\begin{enumerate}
  \item We formulate cross-validation as an estimation problem over a prediction-task distribution, explicitly separating the roles of the validation task generator and the risk estimator. This perspective clarifies that bias in CV arises from mismatches between validation and deployment task distributions and provides a unifying framework for analyzing spatial resampling strategies (Sect.~\ref{sec:deploymentrisk}).
  \item We develop a deployment-oriented validation framework for spatial prediction that combines spatially structured task generators with task descriptors capturing both environmental conditions and task difficulty, in particular prediction distance. Within this framework, we introduce Target-Weighted Cross-Validation (TWCV), which reweights validation losses to align the distribution of validation tasks with that of prediction tasks across the target domain (Sect.~\ref{sec:dwcv}).
  \item Through simulation experiments (Sect.~\ref{sec:simulation} and \ref{sec:virtualspecies}) and a real-world case study in environmental pollution mapping (Sect.~\ref{sec:casestudies}), we demonstrate that (i) task-distribution mismatch is a dominant source of bias in commonly used CV schemes, and (ii) combining spatial task generators with task-aware weighting substantially improves the reliability of performance estimation under realistic sampling designs.
\end{enumerate}

\section{Deployment-Oriented Risk Estimation}
\label{sec:deploymentrisk}

\subsection{Prediction tasks}\label{sec:prediction.tasks}

Consider a spatial prediction problem in which observations
\[
\{(s_i, x_i, z_i)\}_{i=1}^n
\]
are available at spatial locations $s_i$, where $x_i$ denotes a vector of predictors and $z_i$ the observed response variable. A predictive model $f$ is used to generate predictions $\hat z = f(x)$ at locations where observations are not available.

Predictions are typically required across a spatial domain $D$, for example a grid covering a region. Each prediction location induces a \emph{prediction task} that depends on predictors at that location and on the spatial configuration of available training observations, or more generally, descriptors of task difficulty.

We therefore represent prediction tasks by descriptors
\[
T = (x, d),
\]
where $x$ denotes the covariate vector and $d$ summarizes task difficulty.

The inclusion of $d$ reflects that prediction difficulty depends on the configuration of nearby observations relative to the prediction location. In geostatistics, prediction uncertainty increases with distance to observations and sparse sampling \citep{webster.oliver.2007}. Yet even non-spatial machine-learning models can exhibit distance-dependent performance \citep{brenning.2023.ijgis,frank2025spatialrf}.
More generally, $T$ may be any measurable function of the prediction location and training configuration (see Sect.~\ref{sec:discussion.descriptors}).

\subsection{Deployment risk}\label{sec:deployment.risk}

Let $L(y,\hat y)$ denote a loss function measuring prediction error. The predictive performance of a model under deployment conditions can be expressed as
\[
R_{\text{deploy}}(f) =
\mathbb{E}_{T \sim P_{\text{target}}} [L(T,f)],
\]
where $P_{\text{target}}$ denotes the distribution of prediction tasks across the target prediction domain. We assume that this deployment-task distribution is known or can be approximated from a predefined target domain (for example, study area). This formulation formalizes the distinction between validation and deployment conditions introduced in Sect.~\ref{sec:introduction}.

In spatial mapping, this distribution is induced by prediction locations in $D$, their covariate values, and the configuration of training data. The quantity $R_{\text{deploy}}(f)$ thus represents the expected prediction loss across the target domain.

\subsection{Cross-validation as a validation task generator}

In cross-validation (CV), predictive performance is estimated using $n_{\text{val}}$ validation observations from held-out subsets. Let $T_i$ denote the task associated with validation observation $i$, and $L_i$ the corresponding loss. Standard CV estimates predictive risk as
\[
\hat R_{\text{CV}} =
\frac{1}{n_{\text{val}}}
\sum_{i=1}^{n_{\text{val}}} L_i .
\]

This estimator implicitly targets the quantity
\[
R_{\text{CV}} =
\mathbb{E}_{T \sim P_{\text{CV}}} [L(T,f)],
\]
where $P_{\text{CV}}$ denotes the distribution of validation tasks induced by the sampling design and the CV procedure. Conventional CV and other resampling procedures such as its spatial variants or the boostrap can therefore be conceptualized as \emph{task generators} that draw tasks from an underlying task distribution.

When observations are sampled independently from the deployment distribution, $P_{\text{CV}}$ and $P_{\text{target}}$ coincide, and CV is approximately unbiased. Otherwise, differences between these distributions induce bias \citep{quinonero2009dataset,sugiyama2007covariate,Wadoux2021,brenning.2023.ijgis}.

Unbiased estimation of deployment risk requires that validation tasks adequately cover the range of prediction tasks and that appropriate weights align the validation-task distribution with the target distribution, i.e.\ that all prediction tasks encountered during deployment are represented in the validation data. When these conditions hold, weighted CV provides an unbiased estimator of deployment risk \citep{sugiyama2007covariate}.

\subsection{Deployment shift in spatial prediction}

In spatial prediction problems, differences between $P_{\text{CV}}$ and $P_{\text{target}}$ commonly arise from two sources. First, monitoring networks may sample environmental conditions unevenly, leading to differences in the distribution of predictors between sampled locations and prediction locations \citep{Shaddick2024,linnenbrink.2023.nndmcv}. Second, the spatial configuration of training observations relative to prediction locations may differ between validation tasks and deployment tasks, resulting in differences in prediction-distance distributions \citep{brenning.2023.ijgis,mila.et.al.2023}.

Previous work has highlighted the importance of accounting for such structure in validation procedures. Spatial CV strategies attempt to reduce dependence between training and validation samples by enforcing spatial separation \citep{roberts2017,brenning.2012.sperrorest,schratz2024jss,mila.et.al.2023}. However, even when spatial dependence is addressed, the resulting validation tasks may still differ systematically from prediction tasks encountered across the target domain \citep{brenning.2023.ijgis}.

\subsection{Weighted estimators of deployment risk}\label{sec:weighted_estimators}

A natural approach to addressing deployment shift is to reweight validation losses so that validation tasks reproduce the deployment-task distribution. Importance-weighted risk estimation is a standard strategy for addressing dataset shift in machine learning \citep{sugiyama2007covariate}. In the spatial prediction setting considered here, weights can instead be constructed through calibration weighting, which adjusts the empirical distribution of validation tasks to match known margins of the target domain \citep{deville1992calibration,lumley.2004.jss}.

This perspective is closely related to importance-weighted risk estimation under covariate shift \citep{shimodaira2000improving,sugiyama2007covariate}. If the density ratio $w(T) = p_{\text{target}}(T) / p_{\text{CV}}(T)$ were known, deployment risk could be expressed as
\[
R_{\text{deploy}}(f) = \mathbb{E}_{T \sim P_{\text{CV}}}\!\left[w(T)\,L(T,f)\right],
\]
and estimated by a weighted empirical average over validation tasks. In practice, direct estimation of density ratios in high-dimensional task spaces is challenging and often unstable \citep{sugiyama2012density}. Calibration weighting provides an alternative by constructing weights that match selected low-dimensional summaries, implemented here as marginal distributions of discretized task descriptors between validation and deployment tasks \citep{deville1992calibration}, thereby yielding a stable, low-dimensional approximation to importance weighting that directly targets the deployment-task distribution. In this sense, TWCV can be interpreted as a calibration-based importance-weighted estimator of deployment risk.

In the following section we consider two instances of this idea. Distance-weighted CV corrects for differences in prediction-distance distributions, while the more general target-weighted CV estimator balances multiple descriptors of spatial prediction tasks, accounting for both covariate shift and task-difficulty shift.

\section{Distance-Weighted and Target-Weighted Cross-Validation}
\label{sec:dwcv}

Differences in prediction-distance distributions between validation and deployment tasks can induce bias due to distance-dependent prediction error \citep{brenning.2023.ijgis,frank2025spatialrf}.
To address this issue, we consider a weighted estimator that adjusts validation losses according to prediction distance. Let $d_i$ denote the prediction distance associated with validation task $T_i$, defined here as the distance between the prediction location and its nearest training observation within the corresponding CV fold. \emph{Distance-Weighted CV} (DWCV) estimates deployment risk as

\[
\hat R_{\text{DWCV}}
=
\sum_{i=1}^{n_{\text{val}}} w_i^{(d)} L_i ,
\]

where weights $w_i^{(d)}$ are chosen such that the weighted distribution of prediction distances among validation tasks matches the distribution of prediction distances across the target domain.

In practice, prediction distances are discretized into a small number of classes (for example quantile-based intervals) to stabilize the weighting procedure. Let $g_d(T_i)$ denote an indicator vector identifying the distance class of task $T_i$. The weights are then constructed so that

\[
\sum_{i=1}^{n_{\text{val}}} w_i^{(d)} g_d(T_i)
=
\frac{1}{|D|}\sum_{u \in D} g_d(T_u),
\]

where the right-hand side represents the empirical distribution of distance classes among prediction tasks across the target domain.

DWCV addresses mismatch in prediction distance but not differences in environmental conditions between validation and deployment tasks.
These differences may be critical for judging the performance of machine learning models in the deployment situation since challenging prediction tasks with a poor model performance may be more or less prevalent there than on the validation set, as in non-stationary or heteroskedastic environments.

To address both sources of mismatch simultaneously, we introduce \emph{Target-Weighted CV} (TWCV), which balances multiple task descriptors. Let $g(T)$ denote a vector of balancing variables derived from $T=(x,d)$, including discretized prediction distance and environmental predictors.

The deployment-task distribution can be summarized through the empirical margins

\[
m =
\frac{1}{|D|}
\sum_{u \in D} g(T_u),
\]

where the sum runs over prediction tasks induced by locations in the target domain $D$.

TWCV assigns weights $w_i$ to validation tasks such that the weighted distribution of task descriptors matches these target margins. Formally, weights are obtained by solving the calibration equations

\[
\sum_{i=1}^{n_{\text{val}}} w_i g(T_i)
=
m ,
\]

subject to normalization $\sum_{i=1}^{n_{\text{val}}} w_i = 1$.

The resulting estimator of deployment risk is

\[
\hat R_{\text{TWCV}}
=
\sum_{i=1}^{n_{\text{val}}} w_i L_i .
\]

The weights are obtained using \emph{calibration weighting} (also known as raking), which adjusts the empirical distribution of validation tasks to match known margins of the target population \citep{deville1992calibration,lumley.2004.jss}. In this way, TWCV can be interpreted as a calibration-based estimator of deployment risk.

TWCV requires a validation task generator whose support covers that of the deployment-task distribution. We propose buffered resampling with a range of buffer distances as a validation-task generator as it allows us to create relevant combinations of environmental predictors and prediction distances in a more controlled way than unbuffered random CV or leave-one-out (LOO) sampling. Buffered task generation can be implemented as leave-$m$-out resampling with randomly varying buffer radii while ensuring that training sample sizes do not deteriorate too strongly (for example, $\ge 80\,$\% of original sample size).

TWCV therefore provides a deployment-oriented estimator that accounts for both spatial configuration and environmental representativeness of prediction tasks.
It requires a task generator such as buffered resampling that ensures adequate coverage of the deployment task distribution.

\section{Simulation Studies and Case Study}

\subsection{Simulation study}
\label{sec:simulation}

A simulation study was conducted to evaluate the proposed estimators under controlled conditions, assessing how well CV estimators approximate the true deployment risk when validation-task distributions differ from the deployment-task distribution.

We compare conventional non-spatial and spatial estimators as well as distance-weighted and target-weighted CV, evaluating the bias of estimated deployment risk across repeated simulations.

\subsubsection{Data-generating process}

The simulated response surface was generated on the unit square domain
$D=[0,1]^2$ as
\[
Z(s)=\beta_0+\beta_1 x_1(s)+\beta_2 x_2(s)+\varepsilon(s),
\]
where $x_1(s)=s_x$ is a large-scale west--east gradient, $x_2(s)$ is a spatially structured Gaussian field, and $\varepsilon(s) := \sigma_\varepsilon^2(s)\cdot\varepsilon_0(s)$ is a heteroskedastic residual field derived from a mean-zero Gaussian random field $\varepsilon_0(s)$ with exponential covariance $C(h)=\exp(-h/\rho)$ and linearly scaled variance $\sigma_\varepsilon^2(s)=0.4 + 1.2 s_x$, implying a fourfold increase from left to right.

We considered two trend-strength settings $(\beta_1,\beta_2)\in\{(1,0.5),(3,1)\}$ and two residual correlation ranges $\rho\in\{0.05,0.10\}$, with $\sigma_\varepsilon^2=1$.

To assess the effect of irrelevant information on TWCV, we additionally consider $x_3(s)=s_y$ and an independent spatial Gaussian field $x_4(s)$, none of which is included in the above trend.

Simulations were repeated 100 times. For each dataset, a sample of $n=200$ observations was drawn according to one of three designs representing common monitoring configurations:

\begin{enumerate}
\item \textbf{Random sampling}: observation locations were sampled uniformly in continuous space across the domain $D$.
\item \textbf{Clustered sampling}: observation locations were drawn from $10$ spatial clusters.
\item \textbf{Environmentally biased sampling}: observation locations were generated
in continuous space with inclusion probabilities depending on the large-scale
predictor $x_1(s)$, yielding preferential sampling toward one end of the
environmental gradient.
\end{enumerate}

\begin{figure}[t]
\centering
\includegraphics[width=0.9\textwidth]{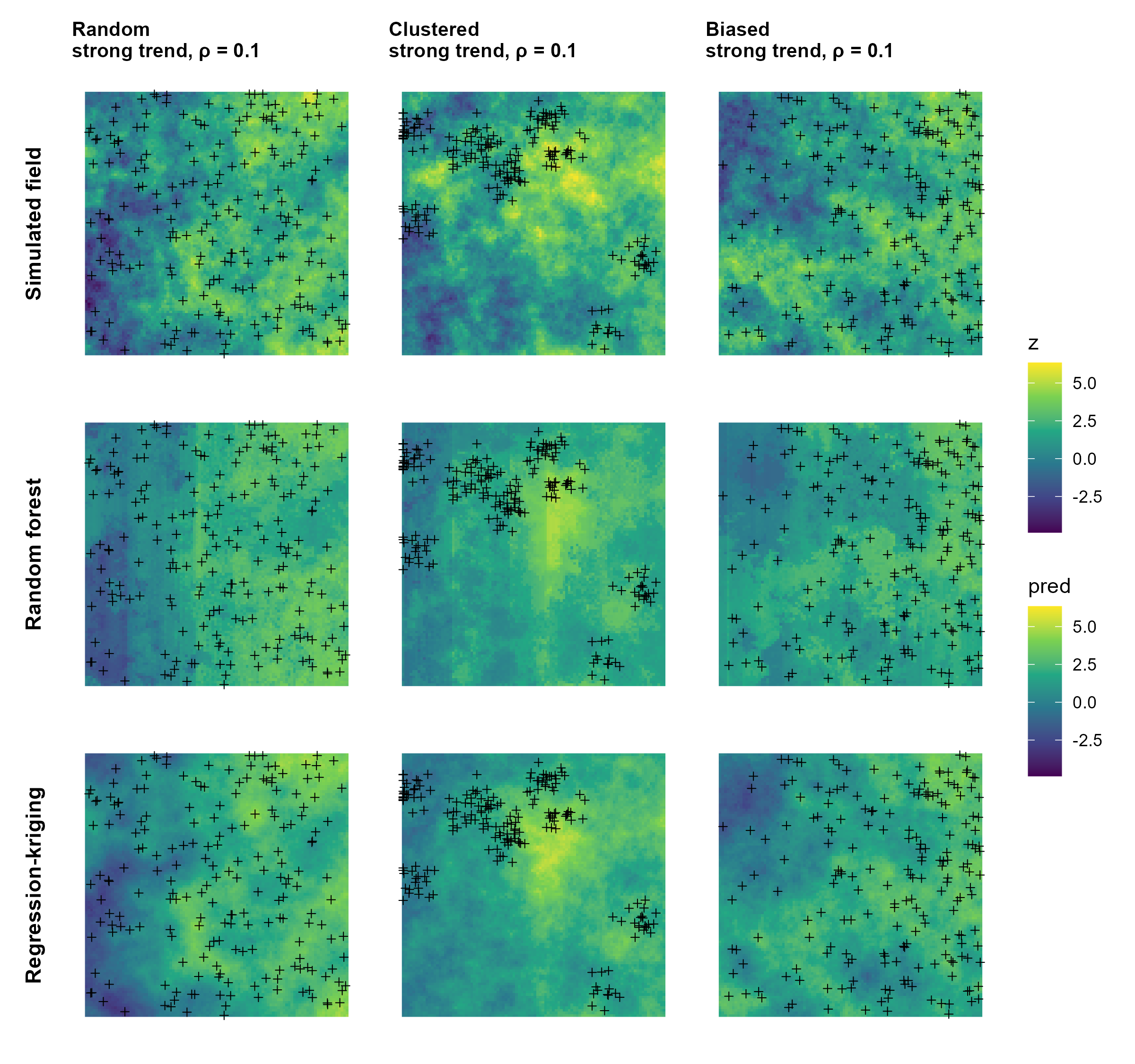}
\caption{
Illustration of the three sampling designs and their effects on spatial prediction. Columns correspond to random, clustered, and preferentially biased sampling for one representative simulation setting (strong trend, $\rho=0.1$). The top row shows the simulated response field on $D=[0,1]^2$ together with sampled locations. The second and third rows show the corresponding prediction maps obtained from random forest and heteroskedastic regression--kriging, respectively.}
\label{fig:sim_sampling_design_prediction_maps}
\end{figure}

\subsubsection{Spatial prediction models}

Two contrasting models were considered: random forest (RF; \citep{breiman.2001.randomforest}) and heteroskedastic regression--kriging (HRK), both using $x_1$ and $x_2$ as predictors.
RF was implemented using ranger \citep{wright2017ranger}. Regression--kriging combines a linear trend with spatially autocorrelated residuals \citep{webster.oliver.2007}.
We implemented RK in gstat \citep{pebesma2004gstat} with a heteroskedastic residual variance model depending on $x_1$. RK is therefore the correctly specified probabilistic model for the simulated situation but requires estimation of trend, variance and semivariogram parameters in each run. Implementation details are provided in Sect.~S2.3 in the Supplement \citep{brenning_suesse_zenodo_twcv}.

\subsubsection{Deployment task and validation design}

Deployment performance was evaluated on a $100\times100$ grid. True deployment risk was computed by comparing predictions to simulated values at these locations.
Prediction tasks on the grid defined the target distribution of task descriptors used for DWCV and TWCV.

Prediction models were evaluated using the following validation procedures.

Conventional CV was implemented as \textbf{LOOCV} and random $k$-fold CV ($k=10$). Spatial CV uses spatially structured folds generated by partitioning observation coordinates into $k=10$ clusters using $k$-means clustering \citep{schratz2024jss}; it was implemented as leave-one-block-out CV (\textbf{LOBOCV}).

We also include $k$-nearest-neighbor distance matching (kNNDM) methods matching prediction-distance (\textbf{Spatial kNNDM}) and feature-space distributions (\textbf{Feature kNNDM}; \citep{linnenbrink.2023.nndmcv}), the latter one being considered experimental.

For \textbf{DWCV}, validation losses were weighted to match prediction-distance distributions. We generated 500 validation tasks using buffered LOO sampling that approximates this distribution.

\textbf{TWCV} extends this approach by balancing multiple descriptors of prediction tasks. Balancing variables were discretized before computing calibration weights. We split each covariate into five categories based on quintiles with a decile-based variant for comparison. We generated 500 buffered LOO tasks to ensure full coverage of the prediction-distance distribution.

In the simulation study, we considered two contrasting TWCV weighting schemes, each with coarse and fine binning. (1) \textbf{TWCV} and \textbf{TWCV--fine} used only the large-scale predictor $x_1(s)$, in addition to prediction distances as in DWCV. (2) \textbf{TWCV--extended} (\textbf{--fine}) and \textbf{Feature kNNDM--extended} additionally used $x_2$, $x_3$ and $x_4$, challenging the calibration-weighting estimator to deal with a multidimensional setting that involves irrelevant information. For the extended and fine variants, we shrunk raking weights towards uniform weights (shrinkage parameter 0.2) to stabilize estimation under the richer balancing specification \citep{lohr2022sampling}. We analyzed weight distributions and effective sample sizes (ESS) to assess estimator stability; details are given in Sect.~S2.1 in the Supplement.

As an additional baseline, we considered importance-weighted CV (\textbf{IWCV}), a covariate-shift approach in which losses are weighted by estimated density ratios \citep{sugiyama2007covariate,shimodaira2000improving}. In contrast to IWCV, which estimates density ratios via probabilistic classification, TWCV constructs weights by enforcing calibration constraints on selected task descriptors.

We implemented \textbf{Random IWCV} and \textbf{Buffered IWCV} using logistic regression with forward selection. Both use the same descriptors as extended TWCV; details are given in Sect.~S2.2 in the Supplement.

For reference, we also considered model-based uncertainty estimates. For HRK, kriging variance provides an unbiased MSE estimate under correct specification \citep{webster.oliver.2007}. For RF, we used the out-of-bag (OOB) error \citep{breiman.2001.randomforest}.

\subsubsection{Evaluation}

For each replicate, true deployment risk was computed as
\[
R_{\text{deploy}}(f)
=
\frac{1}{|D|}
\sum_{u \in D} L(Z(u), \hat Z(u)),
\]
where $D$ denotes the set of prediction locations in the target domain and
$L(\cdot,\cdot)$ is the loss function used throughout the evaluation. In the
experiments presented below we consider squared error loss.

For each estimator, we computed $\hat R$ using the respective validation procedure. Performance was summarized across simulations using RMSE.

\subsection{Virtual species simulation experiment}\label{sec:virtualspecies}

We conducted an additional, ecologically motivated simulation study using the virtual species framework \citep{leroy2016virtualspecies}, which generates spatially explicit species occurrence patterns from 18 environmental predictor variables. We used the implementation and data structure provided by \citet{mila.et.al.2023}, which has been proposed as a benchmark for evaluating spatial CV strategies, but added preferential sampling as an additional scenario not considered in \citet{mila.et.al.2023}. Preferential sampling was applied with respect to one randomly selected predictor; RF and linear regression were used as models. Details are given in Sect. S4 in the Supplement.

\subsection{Case study: air quality mapping}
\label{sec:casestudies}

To illustrate implications of deployment-oriented validation, we consider a spatial prediction problem with non-uniform sampling. The objective is to assess how different CV estimators characterize predictive performance across a deployment region and how task generators approximate the deployment-task distribution.

We analyze annual mean NO$_2$ concentrations across Germany for 2018 \citep{uba_stations_2018,uba_no2_2020} using predictors representing emission sources and environmental conditions. Such models, commonly referred to as land-use regression, are widely used for large-scale air pollution mapping \citep{vizcaino2018lur,hoek2008lur}.
Resulting prediction surfaces are routinely used by environmental agencies for regulatory reporting and public information \citep{kessinger2020airqualityapp,who_no2_2021}. Monitoring networks are typically denser in urban areas and thus biased relative to the full domain \citep{Shaddick2024}.

\begin{figure}[t]
\centering
\includegraphics[width=0.8\textwidth]{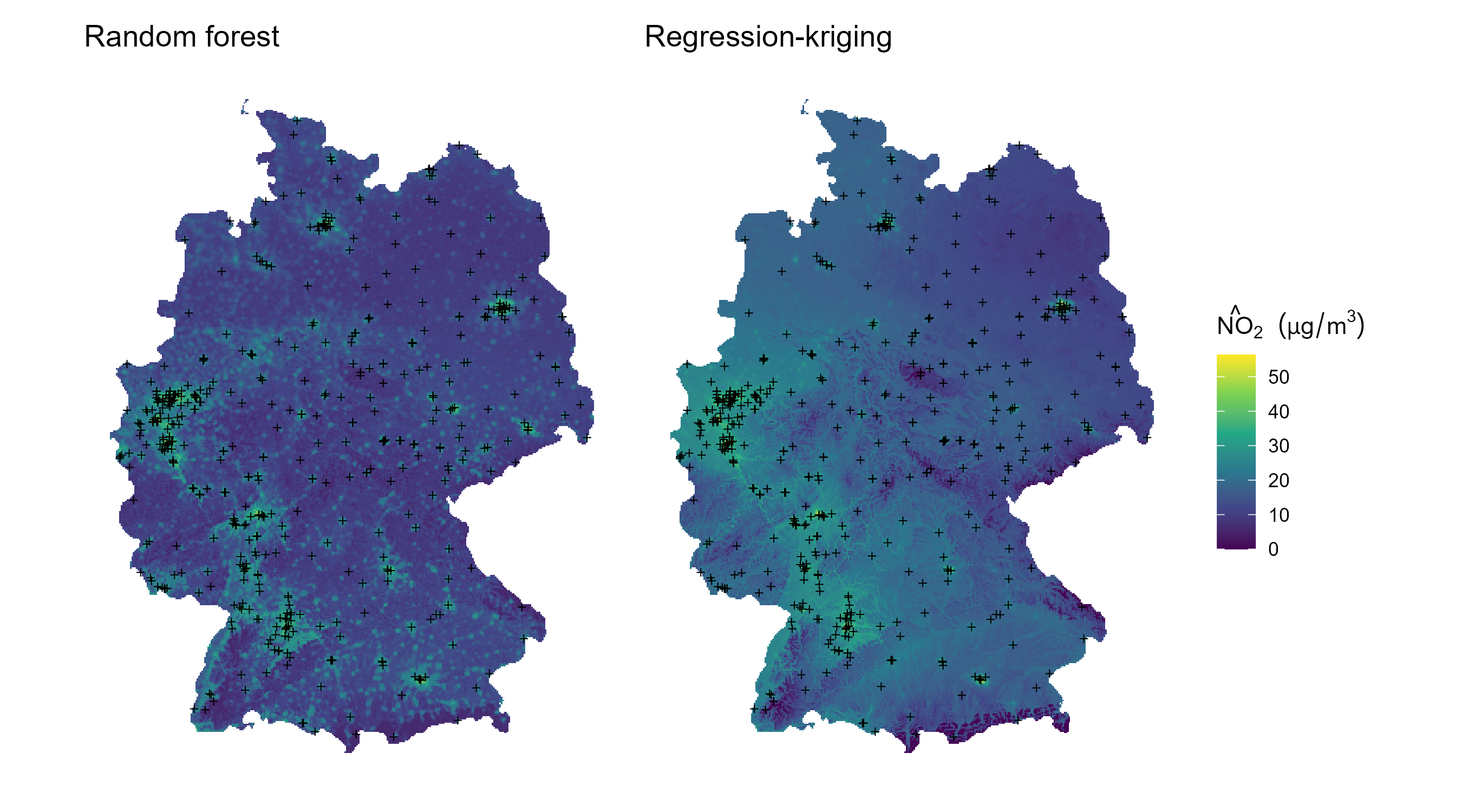}
\caption{Predicted annual mean NO$_2$ concentrations across Germany obtained with random forest (left) and regression--kriging models (right), along with the locations of 503 monitoring stations. Both models use the same set of topographic and demographic covariates.}
\label{fig:no2-prediction-maps}
\end{figure}

The dataset comprises annual mean NO$_2$ concentrations at 503 monitoring stations (Fig.~\ref{fig:no2-prediction-maps}). Predictions were generated on nation-wide 2~km grid. Predictors include elevation, distance to coast, population density, topographic roughness index (TRI), and topographic position index (TPI; data sources: USGS EROS \citep{usgs_gtopo30_2023}, SEDAC \citep{sedac_gpwv4_popdensity_2017}, Copernicus/EEA \citep{eea_corine_clc2018_raster}). Smoothed population density served as a proxy for emission sources.

The monitoring network exhibits a strong bias toward urban areas: $50\,\%$ of the stations are located within the $2\,\%$ most densely populated areas (Fig.~\ref{fig:no2-prediction-maps}). Prediction distances for deployment locations are also substantially larger than the nearest-neighbour distances among monitoring stations.

Exploratory analysis indicated weak but long-range spatial dependence (practical range $\sim 700$~km) and substantial small-scale variability (nugget-to-sill ratio $\sim 0.7$). Linear model residuals are non-stationary and heteroskedastic, with contrasting spatial dependence structures in rural versus urban areas (Fig.~S4 in the Supplement). In particular, urban residuals have a three times higher variance than rural ones.

We used the same task generators and estimators as in the simulation study, excluding extended and fine variants. All predictors were included in TWCV balancing and Feature kNNDM distribution matching. We ran buffered TWCV and DWCV with 10{,}000 validation tasks.
We applied random forests (RF) and homoskedastic regression--kriging (RK). Predictive performance was evaluated using the RMSE.

\section{Results}\label{sec:results}

\subsection{Simulation study}\label{sec:results.simulation}

\begin{figure}[t]
\centering
\includegraphics[width=\textwidth]{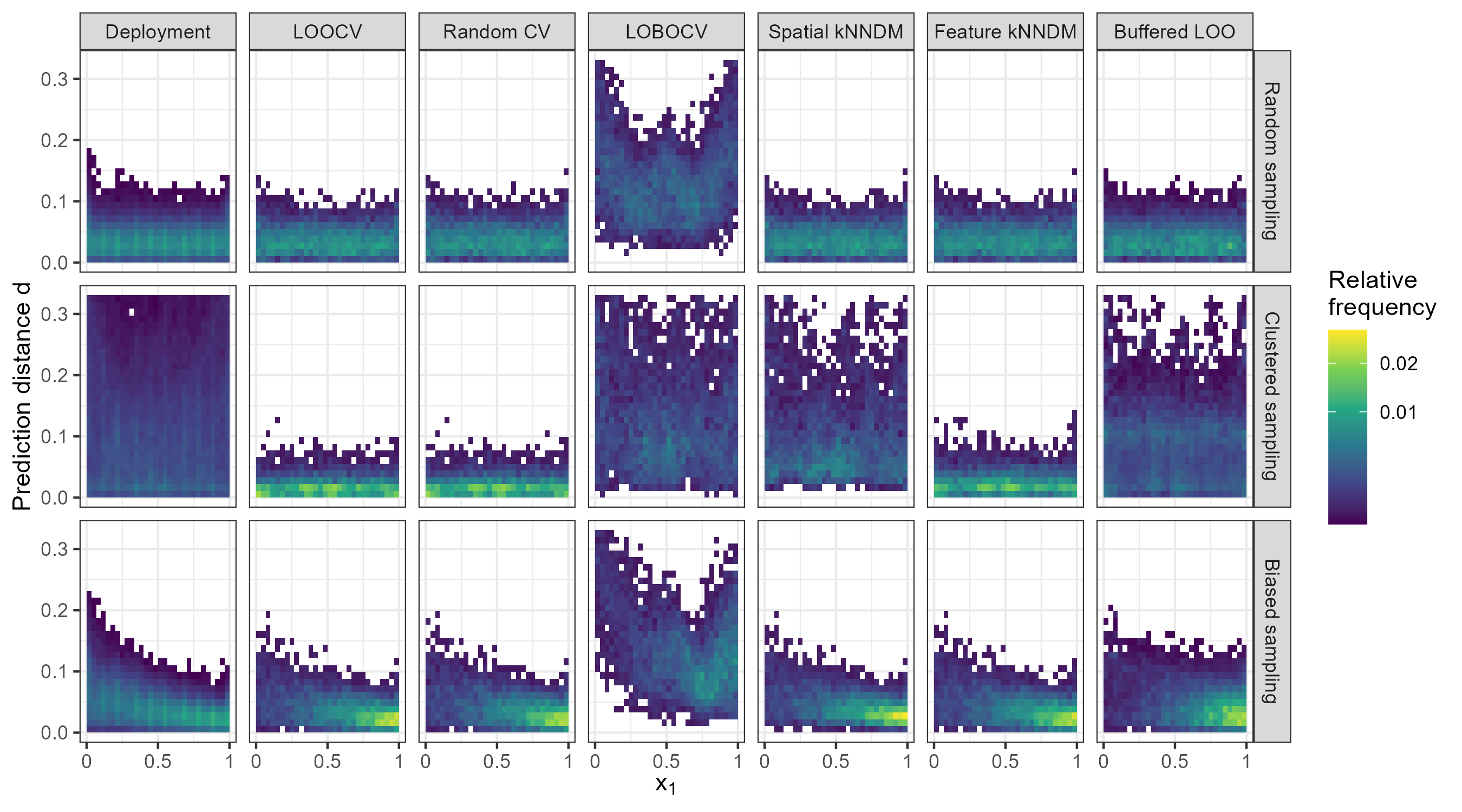}
\caption{
Joint distribution of prediction tasks in $(x_1,d)$ space in the simulation study.
Panels compare the deployment task distribution with validation tasks generated by LOOCV, random $10$-fold CV, LOBOCV, and buffered LOOCV, separately for the three sampling designs.
Here, $x_1$ denotes the large-scale environmental gradient and $d$ the nearest-neighbour prediction distance to the corresponding training set.}
\label{fig:sim_joint_task_distribution}
\end{figure}

Validation task distributions from conventional CV differed markedly from deployment, particularly in prediction distance under clustered and preferential sampling (Fig.~\ref{fig:sim_joint_task_distribution}).
LOOCV and random CV predominantly produced short prediction distances because validation points remained surrounded by nearby training observations. LOBOCV shifted prediction distances towards larger values, though unevenly across the environmental gradient $x_1$. Buffered LOO, in contrast, yielded a broader coverage of the $(x_1,d)$ task space and thus provided an improved coverage of the deployment-task distribution's support. Spatial kNNDM showed slightly greater deviation from the deployment distribution while Feature kNNDM induced task distributions that are closer to random CV.

\begin{figure}[t]
\centering
\includegraphics[width=\textwidth]{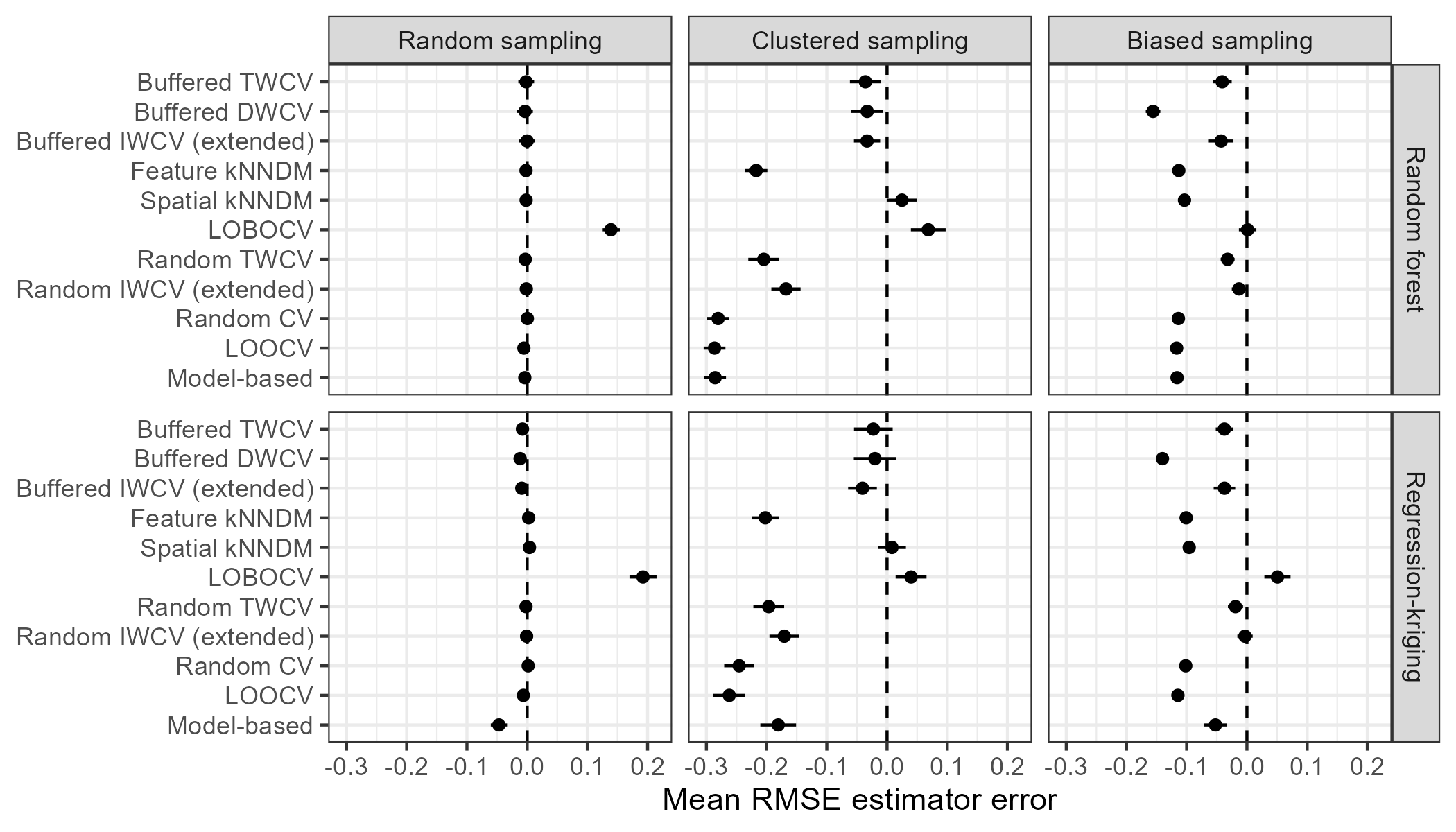}
\caption{
Mean error of RMSE estimators in the simulation study for different validation schemes and model-based uncertainty estimators for one representative simulation scenario (strong trend, $\rho=0.1$).
Points show the mean difference between the estimated RMSE and the true deployment RMSE, with approximate $95\,\%$ confidence intervals.
The dashed line indicates unbiased estimation of deployment RMSE.
Positive values indicate overestimation of prediction error.
Refer to the Supplement for results of other simulation scenarios and estimators.}
\label{fig:sim_mean_rmse_error}
\end{figure}

These differences in task distributions translated into systematic biases of RMSE estimators (Fig.~\ref{fig:sim_mean_rmse_error}). Only buffered TWCV and buffered IWCV remained nearly unbiased across all sampling designs and for both prediction models (RF and HRK). In contrast, non-spatial LOOCV and random CV schemes exhibited substantial optimistic bias under clustered as well as preferential sampling. DWCV and spatial kNNDM removed this bias only in the clustered sampling design, where discrepancies in prediction distance dominate the mismatch between validation and deployment tasks. LOBOCV tended to produce pessimistically biased RMSE estimates, particularly under random and preferential sampling, reflecting its over-representation of harder prediction tasks with large prediction distances. Feature kNNDM was biased in all non-random designs. General patterns were consistent across scenarios, with stronger effects for larger autocorrelation range ($\rho=0.1$ vs.\ $0.05$; Fig.~S1).

Model-based uncertainty estimates showed contrasting behavior for the two models. For RF, the OOB RMSE estimate was optimistically biased under clustered and preferential sampling, consistent with the similarity between OOB and random CV resampling. For HRK, the model-based RMSE derived from kriging variances was nearly unbiased across most scenarios. A small optimistic bias under clustered sampling may be due to semivariogram parameter uncertainties not being propagated into the kriging variances.

Calibration weight diagnostics among buffered DWCV/TWCV variants revealed that under non-random sampling the concentration of weights increased and the effective sample size (ESS) decreased with increasing complexity of balancing scheme, i.e. from DWCV (ESS fraction $80\,\%$ for clustered and $99\,\%$ for biased sampling) and simple TWCV ($58$ and $40\,\%$) to extended TWCV with fine bins ($13$ and $19\,\%$). Under random sampling, where weighting should be redundant, ESS was $>90\,\%$ for DWCV/TWCV and remained at $\sim 80\,\%$ for extended and fine TWCV, but dropped to $50\,\%$ in the ---as expected, over-parameterized--- `extended, fine' variant. Although this was mirrored by increases in extreme weights, TWCV complexity did not affect mean RMSE estimator error (Fig.~S3 in the Supplement).

Overall, the results show that mismatches between validation and deployment task distributions can be an important source of bias in CV-based uncertainty estimation, and that explicitly targeting the deployment-task distribution via TWCV nearly removes this bias when combined with a task generator offering broad enough support.

\subsection{Virtual species simulation experiment}
\label{sec:results.virtualspecies}

Results were consistent with our main findings (Fig.~S7 in the Supplement): Across clustered and preferential sampling biases, only buffered TWCV and buffered IWCV consistently achieved RMSE estimator bias reduction, while especially LOBOCV and plain LOOCV were biased, and Feature kNNDM as well as Spatial kNNDM did not eliminate bias as consistently as the proposed framework. For detailed results and refer to Sect.~S4 in the Supplement.

\subsection{Case study: Air quality mapping}
\label{sec:results.airquality}

\begin{figure}[t]
\centering
\includegraphics[width=\linewidth]{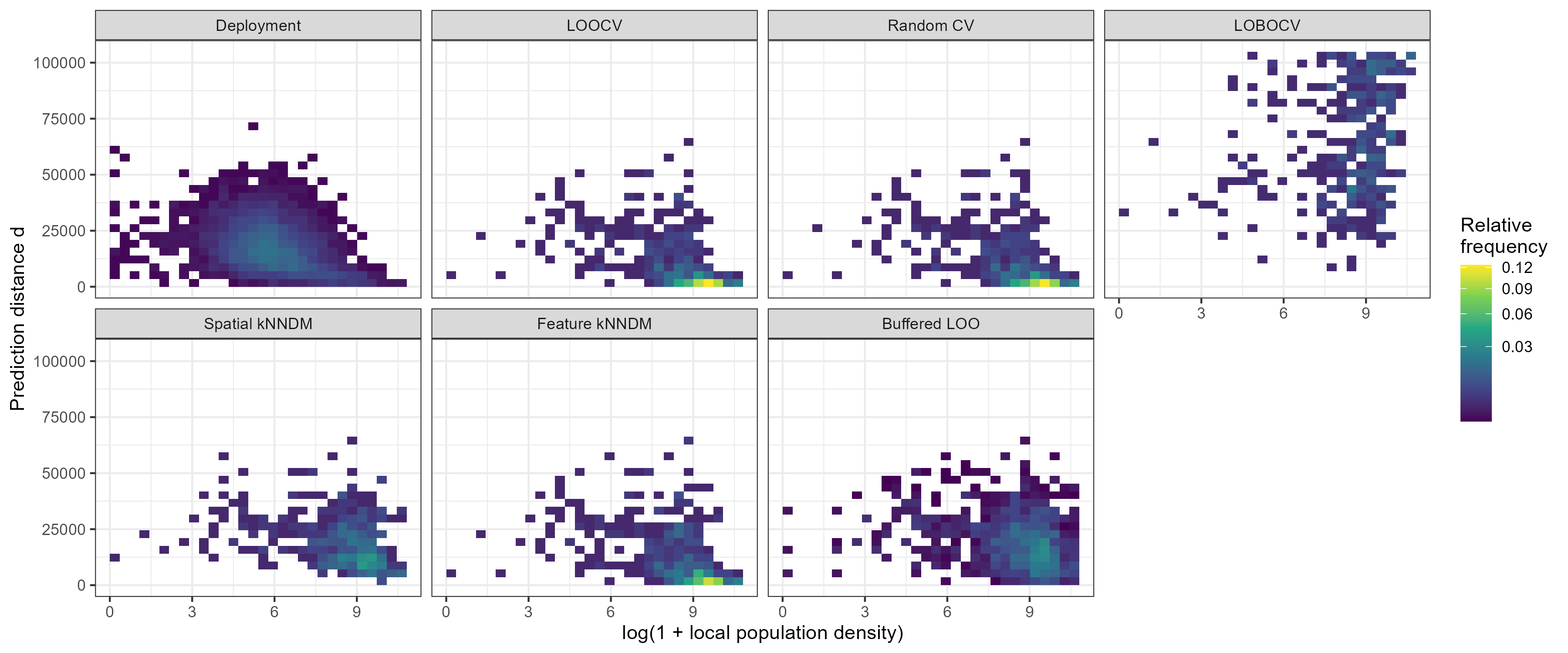}
\caption{Empirical distribution of prediction tasks in the space spanned by population density and prediction distance $d$ for the 2018 NO$_2$ case study in Germany. Panels compare the deployment tasks on the 2~km target grid with validation tasks induced by LOOCV, random CV, LOBOCV, kNNDM, and buffered LOOCV. Local population density is shown on the $\log(1+x)$ scale.}
\label{fig:no2-popd-d-empirical-frequency}
\end{figure}

Among all resampling strategies, buffered LOO provided the most complete coverage of the $(\text{population density},d)$ space (Fig.~\ref{fig:no2-popd-d-empirical-frequency}), providing the variation necessary for weighted estimators. Random CV and LOOCV reproduced the sampling bias toward short prediction distances and high population densities, failing to adequately cover the range of tasks encountered across the deployment domain. LOBOCV overrepresented larger distances while retaining a bias toward high population densities.

\begin{figure}[ht]
\centering
\includegraphics[width=0.8\linewidth]{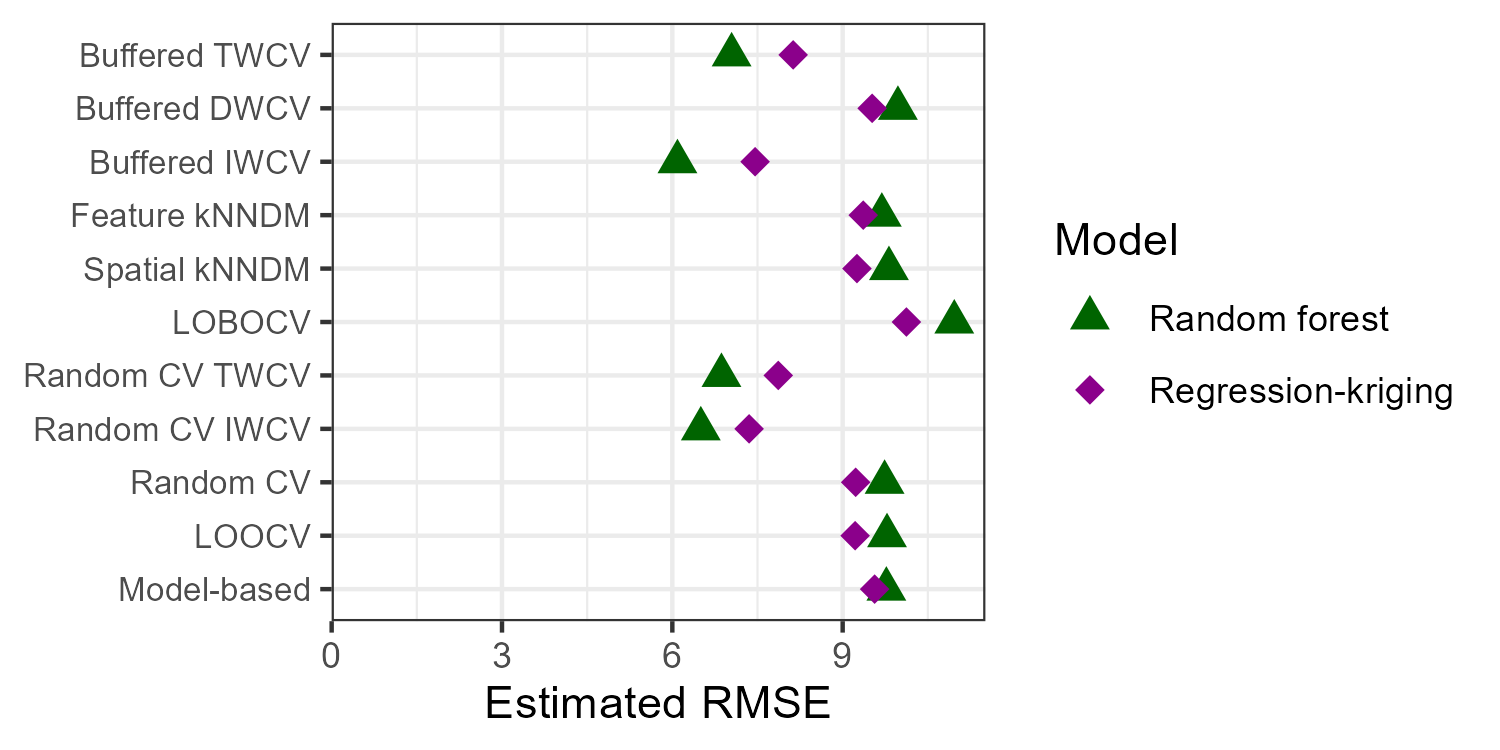}
\caption{Estimated root mean squared prediction error (RMSE) for annual mean NO$_2$ concentrations in Germany obtained from different uncertainty estimators and prediction models. Shown are the model-based estimate, LOOCV, random CV, LOBOCV, and buffered CV in unweighted, importance-weighted (IWCV), distance-weighted (DWCV), and target-weighted (TWCV) form.}
\label{fig:no2-rmse-by-estimator}
\end{figure}

TWCV and IWCV variants yielded RMSE estimates typically $15$--$35\,\%$ lower than other methods, both with buffered LOO and random CV task generators (Fig.~\ref{fig:no2-rmse-by-estimator}). We interpret this as pessimistic bias of unweighted estimators. Because monitoring sites are concentrated in urban areas where NO$_2$ concentrations exhibit higher variability, conventional spatial and non-spatial validation tasks overrepresented urban high-variance conditions relative to the rural-dominated deployment domain. As a result, they overestimated prediction error when averaged over the full prediction domain. This also affected feature-based kNNDM, whose matching was largely influenced by features not affected by covariate shift. Thus, TWCV/IWCV were not optimistic; they rather estimated a defined target, i.e. deployment risk.

Model-based uncertainty estimates from the RK model were pessimistically biased due to the overrepresentation of highly variable urbanized regions in the sample, which affected the model's representation of spatial covariation and prediction variance based on the residual semivariogram (Fig.~S4 in the Supplement). Non-parametric calibration weighting, in contrast, was able to capture such non-stationarities, resulting in more reliable estimators of deployment uncertainty.

Weight-concentration diagnostics revealed that buffered TWCV and IWCV assigned highly unequal weights across monitoring sites, which is an inevitable result of preferential sampling. Although most stations received nonzero weight, the effective sample size of TWCV was only 63 and uncertainty estimates were dominated by a small subset of observations (Fig.~S5). This reflects limited overlap between the validation-task distribution induced by the monitoring network and the target deployment-task distribution.

\section{Discussion}\label{sec:discussion}

This study provides empirical evidence that mismatches between validation and deployment task distributions lead to systematic bias in spatial cross-validation.
These biases arise from the interaction between validation task generation and risk estimation. Spatial resampling addresses the former, while the proposed framework targets the latter through weighting.

\subsection{Task distribution mismatch as a source of bias}\label{sec:discussion.bias}

Building on the task-based formulation introduced in Sect.~\ref{sec:deploymentrisk}, our results show that mismatches between validation and deployment task distributions constitute a primary source of bias in previously proposed spatial and non-spatial CV variants. Different validation schemes induce different task distributions, leading to systematic over- or underestimation of predictive performance depending on how prediction distance and environmental conditions are represented.

While prior work has emphasized dependence between training and validation data as a source of optimistic bias \citep{brenning2005,roberts2017,karasiak.et.al.2021}, our findings indicate that reducing dependence alone is insufficient. Even spatially structured resampling schemes can induce systematic distortions in task characteristics, particularly prediction distance and covariate distributions. These effects are evident in the results in Sect.~\ref{sec:results}, where different validation schemes emphasize different regions of the task space. This observation is consistent with critiques of spatial CV that stress the importance of aligning validation design with prediction conditions \citep{Wadoux2021}.

\subsection{Relation to existing validation strategies}\label{sec:discussion.existing}

Existing approaches to spatial validation primarily address distributional mismatch through resampling design alone \citep{mila.et.al.2023,linnenbrink.2023.nndmcv}. In contrast, TWCV operates at the level of the risk estimator: it relies on task generators that provide sufficient coverage of the task space and adjust for residual mismatch through weighting. These approaches are therefore complementary, representing distribution matching by resampling versus distribution matching by weighting.

Our results suggest that resampling-based matching alone may be insufficient in settings with preferential sampling or multiple sources of shift, whereas weighting provides a flexible adjustment when suitable task descriptors are available. In particular, TWCV can be combined with spatial task generators such as buffered leave-one-out or kNNDM, allowing validation design and risk estimation to be addressed separately. In our experiments, buffered leave-one-out sampling provided better coverage of the task space than random CV, which in some cases failed to cover the support of the deployment distribution.

Recent work on dissimilarity-adaptive CV assigns weights to validation losses based on measures of dissimilarity between training and validation samples \citep{Wang2025}. While this approach also accounts for heterogeneous prediction difficulty, it does not explicitly target a deployment-task distribution but emphasizes difficult prediction tasks. TWCV differs in that it aligns weighted validation tasks with deployment conditions while incorporating task difficulty through explicit descriptors.

Within the broader literature on dataset shift, TWCV can be viewed as a principled approach to correcting such mismatch, and specifically as a calibration-based alternative to importance-weighted risk estimation \citep{shimodaira2000improving,sugiyama2007covariate}. Rather than estimating density ratios directly, TWCV matches marginal distributions of selected task descriptors. This yields a practical estimator in structured prediction settings, where low-dimensional summaries of task distributions are often sufficient to capture the dominant sources of variation.

\subsection{Choice of task descriptors and balancing variables}\label{sec:discussion.descriptors}

The effectiveness of weighted CV depends critically on the choice of task descriptors (Sect.~\ref{sec:prediction.tasks}). In spatial prediction, these descriptors should capture both environmental conditions and spatial configuration.
Prediction distance serves as a key descriptor of task difficulty under spatial dependence \citep{webster.oliver.2007,brenning.2023.ijgis}.

In our experiments, relatively parsimonious sets of balancing variables were sufficient to remove most of the bias. Including many variables can lead to instability due to sparsity in the joint task distribution, a well-known issue in calibration weighting \citep{lumley.2004.jss}. This highlights a trade-off between representativeness and stability in deployment-oriented validation. Similar considerations arise in importance-weighted learning, where flexible density-ratio estimators can capture complex distributional differences but may suffer from high variance in high-dimensional settings \citep{sugiyama2012density}.

\subsection{Relation to extrapolation and feature-space distance}

The proposed framework does not explicitly incorporate measures of extrapolation in feature space. Such measures have been developed to assess model applicability and prediction difficulty beyond geographic proximity \citep{meyer.pebesma.2021.aoa,schumacher.2024.lpd}.

In TWCV, aspects of feature-space mismatch are partially captured through environmental predictors included in the calibration variables, but this does not fully represent local extrapolation in high-dimensional predictor space. Extending the framework to incorporate explicit feature-space similarity measures is therefore a natural direction for future work.

Local point density (LPD) provides one such measure by quantifying the density of training samples in covariate space \citep{schumacher.2024.lpd}. Within the present framework, LPD can be interpreted as a task difficulty descriptor. However, its dependence on model-specific variable weighting makes it less suitable as a general-purpose balancing variable in a model-agnostic validation framework. In this context, LPD is better viewed as a complementary diagnostic tool.

\subsection{Limitations and practical considerations}

Reliable estimation requires sufficient overlap between validation and deployment task distributions. When validation tasks do not adequately cover the support of the deployment distribution, weighting becomes unstable or infeasible. This limitation is analogous to the positivity assumption in causal inference and covariate shift correction \citep{sugiyama2007covariate}. Spatially structured resampling can improve coverage but does not fully resolve this issue under strong sampling bias.

A further requirement is that the deployment-task distribution can be specified or approximated for a given application, for example through a defined prediction domain or a representative set of prediction locations. This assumption may be difficult to satisfy in settings with evolving data-generating conditions, such as non-stationary environmental systems or time-series forecasting \citep{hyndman2018forecasting}, or in applications involving rare extreme events with limited empirical support.

While developed in a spatial context, the framework applies more broadly to settings in which prediction tasks exhibit structured variation. In time-series forecasting, task descriptors may include forecasting horizon or temporal distance to training data \citep{taieb2012review}. In graph-structured data, task difficulty may depend on graph distance or connectivity \citep{chen2022graphdistance}. More generally, the framework can incorporate metadata describing varying reliability of inputs, linking it to domain adaptation settings \citep{quinonero2009dataset,wang2021generalizing}.

\section{Conclusions}\label{sec:conclusion}

We introduced a deployment-oriented validation framework for spatial prediction based on weighted cross-validation. By representing prediction problems in terms of tasks and aligning validation-task distributions with deployment conditions, the framework provides a principled approach to estimating predictive performance under realistic sampling designs. Target-Weighted Cross-Validation (TWCV) implements this idea using calibration weighting and spatially meaningful task descriptors, including prediction distance.

Simulation experiments and a real-world case study demonstrate that mismatches between validation and deployment task distributions are a critical source of bias in commonly used cross-validation strategies. Combining spatially structured task generators with task-aware weighting substantially improves the reliability of performance estimates in such settings.

The proposed framework is readily integrated into existing spatial modeling workflows and applies more broadly to settings with structured variation in prediction tasks. Future work should explore extensions to non-stationary and high-dimensional settings, and assess the framework in additional spatial and non-spatial environmental application domains.

\section*{Code and data availability}

The current version of the code is available on Github at \url{https://github.com/alexanderbrenning/TWCVpaper} under the GNU General Public License v3.0.
The exact version of the code and data used to produce the results and figures presented in this paper is archived on Zenodo under DOI 10.5281/zenodo.20090837 \citep{brenning_suesse_zenodo_twcv}.

\section*{Author contributions}

A.B. conceived the study, developed the implementation,
conducted the formal analysis, and wrote the initial draft of the manuscript.
A.B. and T.S. contributed to the theoretical
development, methodology, and writing of the final manuscript.

\section*{Competing interests}

The authors declare that they have no competing interests.

\section*{Financial support}

This research was supported by the German Environment Agency
(Umweltbundesamt) under grant number 3724237100.

\section*{Acknowledgements}

The authors used ChatGPT (OpenAI) to assist with language editing, restructuring of text, coding, and code documentation. All scientific content, methodological development, analysis, and interpretation were carried out by the authors, who take full responsibility for the final manuscript and associated code.

\clearpage

\bibliographystyle{plainnat}

\bibliography{references}

\end{document}